\newcommand{\todo}[1]{}
\newcommand{\note}[1]{}
\newcommand{\que}[1]{}
\newcommand{\arpit}[1]{}
\newcommand{\atanu}[1]{}
\newcommand{\alex}[1]{}

\PassOptionsToPackage{table,dvipsnames}{xcolor}
\documentclass[acmsmall]{acmart}
\AtBeginDocument{%
  }

\setcopyright{acmlicensed}
\copyrightyear{2018}
\acmYear{2018}
\acmDOI{XXXXXXX.XXXXXXX}

\acmJournal{JACM}
\acmVolume{37}
\acmNumber{4}
\acmArticle{111}
\acmMonth{8}

\usepackage{enumitem}
\usepackage{graphicx}
\setlist[enumerate]{itemsep=8pt}
\usepackage{soul}
\usepackage[most]{tcolorbox}
\usepackage{xcolor}
\usepackage{tikz}

\usepackage{caption}
\newcommand{\paragraphHeadingSpace}{\vspace{4px}}
\newcommand{\bpstart}[1]{
\paragraphHeadingSpace\noindent{\textbf{#1}}%
}

\newcommand{\orangecircleglyph}[2][YellowOrange]{%
  \tikz[baseline=-0.75ex]{
    \node[shape=circle,draw,#1,fill=#1,
      minimum size=9pt, inner sep=0pt,
      text=white, font=\tiny,
      text width=9pt, align=center
    ] (char) {#2};}
}

\newcommand{\greencircleglyph}[2][Green]{%
  \tikz[baseline=-0.75ex]{
    \node[shape=circle,draw,#1,fill=#1,
      minimum size=9pt, inner sep=0pt,
      text=white, font=\tiny,
      text width=9pt, align=center
    ] (char) {#2};}
}

\newcommand{\redcircleglyph}[2][Maroon]{%
  \tikz[baseline=-0.75ex]{
    \node[shape=circle,draw,#1,fill=#1,
      minimum size=9pt, inner sep=0pt,
      text=white, font=\tiny,
      text width=9pt, align=center
    ] (char) {#2};}
}

\tcbset{
  customsidebar/.style={
    colback=blue!10!white,      
    colframe=blue!50!white,     
    coltitle=white,              
    fonttitle=\bfseries,
    boxrule=0pt,
    arc=0pt,
    left=0pt,
    right=0pt,
    top=0pt,
    bottom=0pt,
    width=\linewidth,
    sharp corners,
    enhanced,
  }
}

\newtcolorbox{customsidebar}[2][]{
  customsidebar,
  title=#2,
  #1
}

\begin{document}

\title[Agentic Enterprise]{Agentic Enterprise: AI-Centric~User to User-Centric~AI}

\author{Arpit Narechania}
\email{arpit@ust.hk}
\orcid{0000-0001-6980-3686}
\affiliation{%
  \institution{The Hong Kong University of Science and Technology}
  \country{Hong Kong SAR, China}
}

\author{Alex Endert}
\affiliation{%
  \institution{Georgia Institute of Technology}
  \city{Atlanta}
  \country{USA}}
\email{endert@gatech.edu}
\orcid{0000-0002-6914-610X}

\author{Atanu R Sinha}
\affiliation{%
  \institution{Adobe Research}
  \city{Bengaluru}
  \country{India}}
\email{atr@adobe.com}
\orcid{0000-0002-7949-7982}
\renewcommand{\shortauthors}{Narechania, Endert, and Sinha}

\begin{abstract}
After a very long winter, the Artificial Intelligence (AI) spring is here. Or, so it seems over the last three years. AI has the potential to impact many areas of human life--personal, social, health, education, professional. In this paper, we take a closer look at the potential of AI for Enterprises, where decision-making plays a crucial and repeated role across functions, tasks, and operations. We consider Agents imbued with AI as means to increase decision-productivity of enterprises. We highlight six \emph{tenets} for Agentic success in enterprises, by drawing attention to what the current, \texttt{AI-Centric~User} paradigm misses, in the face of persistent needs of and usefulness for \emph{Enterprise Decision-Making}. In underscoring a shift to \textbf{\texttt{User-Centric~AI}}, we offer six tenets and promote market mechanisms for platforms, aligning the design of AI and its delivery by Agents to the cause of enterprise users.
\end{abstract}



\received{20 February 2007}
\received[revised]{12 March 2009}
\received[accepted]{5 June 2009}

\maketitle

\begin{tikzpicture}[remember picture,overlay]
  \node at ([yshift=-1cm]current page.north) {\footnotesize Preprint — not for distribution.};
\end{tikzpicture}

\section{Introduction}
The promise of Artificial Intelligence (AI) has been prognosticated for decades. The pace has decidedly picked up over the last three years with apparent versatility of transformers~\cite{vaswani2017attention} and diffusion models~\cite{ho2020denoising} for text and image based data, respectively. There has been a proliferation of AI Assistants--such as chatbots and co-pilots--in personal and professional domains, albeit yielding mixed results for enterprises\footnote{https://www.mckinsey.com/capabilities/quantumblack/our-insights/seizing-the-agentic-ai-advantage}. At the same time, discussions about Agents~\cite{huang2024position, durante2024agent, shavit2023practices}
have provided a fillip for AI. 
However, lost in all this is the headwind coming from the current, widely practiced regime of \textbf{\texttt{AI-Centric~User}}, whereby humans do the heavy lift of adjusting to inflexible AI. For example, (i) a typical user must learn to post queries in a specific manner to get appropriate and useful responses from AI~\cite{google2024prompting101};
or, (ii) AI models, aligned with general purpose user preferences, miss out on the needs for specific purpose, or for processes and workflows, or of specialists and experienced users~\cite{shi2023specialist}. Moreover, most of the development focus of enterprises is on the AI itself, less so on the expectations of users.

To wit, we refer to \texttt{AI} as the capability of a tool and the tool itself, including but not limited to generative AI (GenAI). We posit that to realize the promise of AI, a shift away from the paradigm of \texttt{AI-Centric~User} to models of \textbf{\texttt{User-Centric~AI}} is needed, wherein such AI adapt according to users, their specific tasks, workflows, and processes. For enterprises, we assert that the shift is attainable through task-specific  Agents and their efficient and effective organization. 
We refer to \texttt{Agents} as systems that independently accomplish tasks on a user's behalf~\cite{openai2023practical}.
Hitherto, the operational lens of industry views Agents through alternative agentic frameworks that pivot around the paradigm of \texttt{AI-Centric~User}~\cite{openai2023practical,google2025agentspace, anthropic2024multiagent}. In this paper, we take a step back and view the space of AI and its delivery via Agents through the lens of users and beneficiaries (hereafter, users) with focus on \textbf{Enterprise Decision-Making}, an important testbed and a milestone for progress in AI and agentic thinking. We offer technological success criteria for efficient and effective impact of AI on enterprises, toward a vision of \texttt{User-Centric~AI}.

\begin{customsidebar}{Embracing \texttt{User-Centric~AI} in Enterprise Decision-Making}
  Our labeling of current efforts around GenAI as \textbf{\texttt{AI-Centric~User}} stems from the heavy lift enterprise users do trying to extract ever more from GenAI, \textit{as if}, it is the ultimate repository, whose yield is limited only by the incapability of users. Our proposed \textbf{\texttt{User-Centric~AI}} shifts the burden to AI itself, which must incorporate user's agency, their expectations, and also their approach to tasks. Enterprise users remain accountable for their tasks and sub-tasks that contribute to decision-making, whether strategic or tactical. In addition, these users' tasks and decisions often have different dependencies with and accountability to other users. Subsuming these user-centric concepts in AI is an important step to attain its potential for \emph{Enterprise Decision-Making}.
  \end{customsidebar}

\bpstart{Enterprises, their Users, and Decision-Making.} The long term viability of AI and its delivery via Agents rides on capital expenditure by enterprises\footnote{\url{https://a16z.com/ai-enterprise-2025/}}\footnote{\url{https://www.mckinsey.com/industries/technology-media-and-telecommunications/our-insights/the-cost-of-compute-a-7-trillion-dollar-race-to-scale-data-centers}}, yet ``only 1 percent of company executives describe their GenAI rollouts as ``mature''\footnote{\url{https://www.mckinsey.com/capabilities/quantumblack/our-insights/the-state-of-ai}}.'' As well, ``despite the endless announcements about how firms are ushering AI into their operations, few make much use of the technology for serious work\footnote{\url{https://www.economist.com/finance-and-economics/2025/05/26/why-ai-hasnt-taken-your-job}}.'' The success of AI depends on the nature of expectations and \textit{net} benefits that may accrue from enterprises' AI usage, which notably is qualitatively different from AI usage in the personal domain. Enterprise users are accountable to someone else, making their behaviors different from their individual self in personal situations. For example, many users may participate in a task; a user's task and solution may be customized; value of a tool is determined by repeated usage; the demand for a tool has enterprise-specific dependencies on tasks and tools that others use; 
and so on\footnote{\url{https://hbsp.harvard.edu/product/8145-PDF-ENG}}. All this makes development of AI for enterprises challenging; but with a large upside if the challenges can be overcome\footnote{\url{https://www.mckinsey.com/capabilities/quantumblack/our-insights/seizing-the-agentic-ai-advantage}}. Agents can help, only if AI overcomes the lack of a deep understanding of enterprise users. 
For enterprises, we particularly drill down on \emph{decision-making} due to its omnipresent role, its ambiguity and uncertainty, its time-consuming nature, and wider opportunity for impact-of-AI, by moving from human decision-making to more automated decision-making. For AI, understanding enterprise decision-making allows a discrete change from low-value-added AI Assistants to high-value added Agents. We identify six \emph{tenets} that can be infused in \texttt{User-Centric~AI}, and contribute substantively toward AI and organization of Agents into \texttt{Platforms}. Our vision complements existing principles, practices, and lessons for successful adoption of AI in enterprises~\cite{openai2024enterpriseAI, anthropic2024multiagent, satav2025enterprise}. 

Notably, prior work on human-centered AI~\cite{shneiderman2022human} and human-AI collaboration~\cite{wang2020human, shao2025future}, predating large language models (LLMs) and GenAI, emphasize human values, ethics, and usability. In this paper, we contribute by carving out such conceptual gaps in current AI research.
We particularly draw attention to the underexplored, albeit central role of users, their agency, and their contribution to the ecosystems. We raise questions around the hidden labor of users who not only utilize AI, but also contribute to it either directly during training or by providing explicit feedback to AI responses, or indirectly through providing their data passively while using a system. Unlike general-purpose AI frameworks, we advocate deep integration of AI with users' beliefs, expectations, tasks, workflows and decision-making contexts. Additionally, in a user's \texttt{Workflow} involving AI, we try to foresee points of commonality and conflict between \texttt{Users} and \texttt{Agents}, and propose coordination-aware, market-mechanism-oriented \texttt{Platforms} of specialized AI agents to foster commonality and diffuse conflict. This shift represents a major rethinking of Agents as incentive-compatible and responsive partners which can even improve the user cognitively. 
Lastly, with traditional models proven in many analytic tasks for enterprises, it is necessary to offer flexibility in AI to complement those. Thus, the unique challenges and opportunities of AI in a user-centric paradigm may require thinking beyond GenAI and Large Language Models (LLMs).

\begin{customsidebar}{Six Tenets of \texttt{User-Centric~AI} in Enterprises}
  \begin{enumerate}[nosep, left=0cm, labelsep=0.2cm]
    \item [1.] Primacy of Process-Orientation in AI.
    \item [2.] Forward Thinking AI.
    \item [3.] Locally Privacy Preserving Agent-Learning.
    \item [4.] Market Mechanism Platform for Agents and Users.
    \item [5.] Agent Risk-Reward Diversity and Quality-Price Diversity.
    \item [6.] Low Entry, Exit Barriers for Agents.
  \end{enumerate}
\end{customsidebar}

\section{The Premise}\label{sec:premise}

As part of our premise, we consider three entities--\texttt{\textbf{Users}}, \texttt{\textbf{Agents}}, \texttt{\textbf{Platforms}}--and a \texttt{\textbf{Workflow}}. Below, we define and clearly delineate a set of assumptions to be used as building blocks for the rest of the paper. These assumptions work as boundary conditions for our propositions.

\bpstart{\texttt{Users}}: A group which may benefit from AI. They could be performing tasks for an enterprise collectively, or, individually. We make the following assumptions about users:

\begin{itemize}[left=0.75cm, labelsep=0.27cm]
    \item [\textbf{UA1}] \emph{User's Agency}: Users seek agency and vary in skills from other users, and across tasks.
    \item [\textbf{UA2}] \emph{Discoverability}: Users self-select whether to discover AI tools and not the other way round.
    \item [\textbf{UA3}] \emph{User Heterogeneity}: For a given task, users are diverse in expectations and utilities from AI.
    \item [\textbf{UA4}] \emph{Task Heterogeneity}: Across tasks, a user has diverse expectations and utilities from AI.
    \item [\textbf{UA5}] \emph{Utility}: Users aim to maximize their own utility from AI.
    \item [\textbf{UA6}] \emph{Privacy}: Users vary in expectations about degrees of protection of information they share.
\end{itemize}

\bpstart{\texttt{Agents}}: ``Agents are systems that independently accomplish tasks on your behalf''~\cite{openai2023practical} by using a language model to manage workflows, make decisions, correct errors, and dynamically interact with external tools. 
Essentially, they are entities through which AI is delivered to users. These could be a single entity or a collection of entities. We make the following assumptions about agents:

\begin{itemize}[left=0.75cm, labelsep=0.27cm]
    \item [\textbf{AA1}] \emph{Agent Manifestation}: An agent can be a user-assisting tool or an autonomous operator.
    \item [\textbf{AA2}] \emph{Task Specialization}: An agent is specialized for a task.
    \item [\textbf{AA3}] \emph{Degree of Autonomy}: An agent can be controlled or autonomous.
    \item [\textbf{AA4}] \emph{Communication Within}: Autonomous agents may communicate directly with other agents.
    \item [\textbf{AA5}] \emph{Communication Outside}: Autonomous agents may communicate directly with users.
    \item [\textbf{AA6}] \emph{Rewards}: Agents are endowed with varying rewards by the platform.
    \item [\textbf{AA7}] \emph{Types of Models}: Agents can represent GenAI or traditional models (we are agnostic to the type) and depend on user demand and supply from the agent organization and platform.
    \item [\textbf{AA8}] \emph{Disclosure}: Agents communicate their capabilities publicly, whether truthful or not.
\end{itemize}

\bpstart{\texttt{Platform}}: A planner, which is an entity in charge of delivering benefits of AI to users. It can be third-party or enterprise-governed. We make the following assumptions about the planner:

\begin{itemize}[left=0.75cm, labelsep=0.27cm]
    \item [\textbf{PA1}] \emph{Accountability}: The planner is accountable to users for delivering AI.
    \item [\textbf{PA2}] \emph{Organization}: The planner uses a set of agents.
    \item [\textbf{PA3}] \emph{Discoverability}: Planner needs to discover capabilities of agents.
    \item [\textbf{PA4}] \emph{Management}: The planner manages the agents to maximize own utility.
    \item [\textbf{PA5}] \emph{Communication}: Users communicate with the planner, which communicates with agents. Planner can use a confederate, or orchestrator, which is then the conduit for communication.
    \item [\textbf{PA6}] \emph{Safety}: The planner is responsible for degrees of safety in outputs. The degree depends on the task and user's stated appetite for risk.
\end{itemize}

\bpstart{\texttt{Workflow}}: ``A workflow is a sequence of steps that must be executed to meet the user's goal''~\cite{openai2023practical}.
For expositional ease, consider users who perform data analytics as part of their work for an enterprise--small, medium or large.  We define the following simplified workflow--comprising four discrete tasks common in an enterprise setting--as a \textit{running example} for the rest of the paper. This workflow is not meant to be exhaustive, but only an illustration.

\vspace{0.2cm}

\begin{enumerate}[nosep, left=0.15cm]
    \item [1.] \textbf{Data} (Preparation)
    \begin{enumerate}[nosep, left=0.15cm]
        \item [$\rightarrow$ 2.] \textbf{Model} (Selection, Application)
        \begin{enumerate}[nosep, left=0cm]
            \item [$\rightarrow$ 3.] \textbf{Results} (Evaluation, Verification)
            \begin{enumerate}[nosep, left=0.15cm]
                \item [$\rightarrow$ 4.] \textbf{Presentation} (Inference, Recommendation)
            \end{enumerate}
        \end{enumerate}
    \end{enumerate}
\end{enumerate}

\vspace{0.2cm}

Each of the four tasks has several sub-tasks and lower-level analytic decisions contained within it. For example, the model selection step consists of gathering a collection of models, ranking or filtering them based on task criteria, and deciding on either a single model or an ensemble approach. Having such discrete tasks makes it easier to assign an agent to it. Agents can also collaborate with each other on a task.  
Agents can also plan~\cite{wei2025plangenllms, huang2024understanding} and reason~\cite{ferrag2025llm, bandyopadhyay2025thinking} within this workflow to the extent desired.
Lastly, we let users freely choose agents to fit their workflow. For instance, one user might only need help with model selection, while another might run a model, evaluate it, and then return to data preparation to create more features.

\vspace{0.25cm}

\section{Tenets of \texttt{User-Centric~AI}}
Under the premise in Section~\ref{sec:premise}, we propose six tenets to guide AI toward an agentic enterprise. First, we identify three necessary primitives for AI to recognize and endow agents with. Then, we present the tenets as success factors for agentic platforms serving enterprise users.

\subsection{Primitives}

We first present three user-first primitives, emphasizing user-centrality.
\begin{itemize}[left=0.35cm]
    \item [I.] User Agency.
    \item [II.] Agent Foresight.
    \item [III.] User Feedback and Agent Learning.
\end{itemize}

In touting (I), drawing from formidable research in psychology~\cite{bandura2006toward}, we recognize the importance of incorporating the following user level criteria: (a) Users' expectations and beliefs, since these impact the realization of their experiences of interactions with AI~\cite{narechania2025guidance, lapple2019learning};
(b) Users' tasks often involve subjectivity, with no single correct answer; only the user fully understands their task and the potential benefits from AI, making their \textit{ex post} evaluation as important as any \textit{ex ante} assessment of AI;
(c) Users are heterogeneous in tasks performed suggesting differences in expectations and beliefs across tasks; (d) Users are heterogeneous in skill levels for a given task, suggesting differences in expectations and beliefs for the task as well as benefits sought from the task; (e) Users want to experiment with AI interactions for their own workflow~\cite{hbr2023newhumanmachine}, which only they know by (b), suggesting expectation of flexibility and experimentation in tools. 

In advocating (II), and to incorporate (a)--(d) above, AI needs to be imbued with Foresight, which we define as \emph{``Looking ahead beyond the immediate next response, suggesting a degree of anticipation of user's reaction to its own response and interjecting for clarifications \textit{judiciously} and with alternatives.''} Note that we do not reproduce other characteristics required of AI which are espoused in the extant literature~\cite{acharya2025agentic,hughes2025ai}. We only highlight Foresight in addition to those.

In emphasizing (III), and to complement (I) and (III), incorporation of heterogeneous feedback for a task and for heterogeneous tasks is necessary for AI to learn to meet and adapt to heterogeneous users' expectations and beliefs. Without an infrastructure for multi-user, multi-task, multi-agent continuous feedback and learning, along with recognition of heterogeneities in users and tasks, no framework can approach a degree of completeness. 
For example, incompleteness in user's prompts about tasks can be anticipated to build into the agents' behaviors. These are especially pertinent with the talk about AI having turned toward agents and agentic frameworks.

\subsection{Tenets}
Based on the primitives, we present six tenets as guidelines for enterprises and platforms, who serve the former. Some apply broadly to enterprise users, others describe desired agent traits, and some target platforms. Not all tenets are needed in every case; a subset may be enough. Each tenet is stated with its rationale and illustrated by an example from the \texttt{Workflow} (introduced in Section~\ref{sec:premise}).

\subsubsection{Tenet 1: Primacy of Process-Orientation in AI}\hfill

\noindent\underline{Rationale}: Broadly, value to users is of two types: (i) Outcome, and (ii) Process. Given UA1, UA3, and UA4, the sample space of outcomes is not known, and thus, AI cannot meet all users' expectations on outcome-quality. Instead, steps or tasks of processes can be offered with respect to workflows, where different users call in at different tasks in the workflow as and when needed (UA4, AA7, PA2). Giving users easy-to-assemble tools to build their own workflows is beneficial. This requires AI to have a deeper understanding of users' workflows and learn how to improve their efficiency. In other words, the focus shifts from training models on outcome data to training on process data. This process-oriented approach shifts responsibility for outcome quality from the platform to the users, and \emph{when focused on process, users retain agency}~\cite{shao2025future, krakowski2025human, heer2019agency} (UA1, UA5).
For instance, users can satisfy their agency through experimentation, trial and error, and achieve more self-actualization. The platform reduces effort on excessive focus on outcome-quality, which is difficult to satisfy under the various heterogeneities (UA3, UA4, PA6). 

\noindent\underline{Example}: 
An outcome orientation focuses on task specific response regardless of the user's overall need. An experienced enterprise user may need help with model selection and running, but does not need help with other tasks. This user knows that model selection depends on both the task prior to it, data preparation step, and the tasks the user wants after it; say, what type of presentation the user wants. An outcome orientation is inadequate since it does not consider the user's presentation need. A process orientation by AI allows users to plug in a task within their workflow, which knows about the presentation need and can inform them about model selection.

\subsubsection{Tenet 2: Forward Thinking AI}\hfill

\noindent\underline{Rationale}: Enterprise users have diverse expectations and goals across tasks, many involving long-term planning and complex workflows (UA3--5, AA2, PA2). Reactive AI platforms that respond only to current user inputs lack the foresight needed for meaningful strategic support. This limits AI's ability to help users anticipate risks, plan ahead, or manage multi-step processes, leading to fragmented and short-sighted assistance. Shifting to Forward Thinking AI that proactively forecasts needs, warns of potential errors early, and guides users forward empowers better decisions and broader goals (UA5, AA4--5, PA3).

\noindent\underline{Example}: Forward Thinking AI within a workflow can occur for a task and for the workflow. In response to a user's ask about model selection, the AI can give the best response. A forward-looking AI can determine additional follow-up questions from the user without her asking and frame the response accordingly. It can offer the user alternative models with pros and cons rationale with respect to her expected follow-up questions, which informs user's selection. Furthermore, knowing the user's workflow a Forward Thinking AI can consider her presentation need as well.

\subsubsection{Tenet 3: Locally Privacy Preserving Agent-Learning}\hfill

\noindent\underline{Rationale}: AI platforms that treat user data uniformly and rely on one-size-fits-all agent learning fail to meet the diverse skills and varied expectations of users across tasks (UA1, UA3--5, AA6--7, PA1). While generalist agents work well for novices, experts benefit from integrating their specialized knowledge to improve engagement and drive continuous agent improvement. Differentiating data and feedback by user expertise and privacy preferences enables personalization and respects user agency (UA1, UA5--6, AA7), encouraging experts to share valuable knowledge without risking privacy or competitive concerns (AA7--AA8). To meet enterprise needs, platforms (PA1, PA3, PA6) can use partitioned or federated learning that tags data by expertise and privacy, supports user-specific agent training, lets users control AI learning contributions, and applies privacy safeguards with incentives for expert participation. Without expert input, platform appeal drops. This privacy-focused, on-demand approach builds reputation, improves personalization, and sustains a dynamic ecosystem where novices and experts drive ongoing innovation and growth.

\noindent\underline{Example}: Consider the data preparation step of our running example \texttt{Workflow}. While prior art exists for feature selection from data, creating the feature that is most useful for a modeling task is as much an art and depends on user experience and judgment. As any expert analyst knows, a feature is important not only based on its contribution to model accuracy, but also on its actionability and managerial meaningfulness. Those two latter aspects are private information to an expert who may not want to share even with all users in the same enterprise in order to preserve their own worth. If the AI Agent gleans and learns this expertise from the user, it can percolate to others, going against the wishes of the user. Anticipating this, a user may not want to interact with the Agent AI, reducing its learnability. Giving the user control of what enters into the AI's learning maintains locally privacy preserving agent learning. In other words, an AI can `think global, act local.'

\subsubsection{Tenet 4: Market Mechanism Platform for Agents and Users}\hfill

\noindent\underline{Rationale}: Given users' diverse preferences and risk tolerances (UA1--5), and the uncertainty in AI and agent quality relative to user needs (AA2, AA6, PA4), we propose a market mechanism~\cite{akerlof1978market,myerson2008perspectives}. It features specialized agents offering varied skills, behaviors, and risk-reward profiles, enabling users to choose agents that best fit their needs, promoting ownership and flexibility in AI outputs (UA1, UA3, AA2, AA6, PA2, PA4). From the platform perspective, a decentralized agent organization reduces the risk of reputation damage from any single poor-performing agent, boosting resilience and encouraging innovation through competition and collaboration (AA6, PA4--5). While centralized control eases management and ensures consistency, it limits user agency and adaptability (UA1, PA4). Organizing agents as interoperable, loosely coupled entities in a market balances user autonomy with quality and reputation safeguards, fostering experimentation, customization, and strong process integration (UA1, UA5, AA2, PA2, PA4--5). Decentralized does not mean there are no checks and balances on ``underperforming'' agents; it only means the planner's role shifts to managing agents through incentives to ensure user success rather than direct control (PA4, AA6).

\noindent\underline{Example}: 
Each task-agent discloses its capabilities publicly so that users know what each can and cannot do; each agent also receives feedback from users after task completion. All this renders incentive compatible reward to the agent creator / developer to improve the agent. Without a market mechanism such disclosure and feedback are rendered less effective.  

\subsubsection{Tenet 5: Agent Risk-Reward Diversity and Quality-Price Diversity}\hfill

\noindent\underline{Rationale}: Diversity in agent offerings is necessary to address the heterogeneity of user tasks (UA4, AA2, AA6). Given the heterogeneity in user expectations and utilities across tasks and the principle that users aim to maximize their own utility (UA5), AI platforms should allow agents to self-select their behavior styles and risk-reward profiles (UA1, AA2, AA6, PA6). Since agents cannot fully eliminate undesirable behaviors due to task heterogeneity, supporting diverse agent behaviors enables users to match risk profiles to their preferences (UA5, AA5--6, PA6). Users comfortable with high-risk, high-reward outcomes can choose more exploratory agents, while others may prefer safer, conservative ones. This flexibility preserves user agency and fosters innovation, while the platform manages a diverse set of specialized agents to meet varying user needs, creating a dynamic ecosystem that balances creativity and safety (UA1, UA5, AA2, PA2, PA6).

\noindent\underline{Example}: Consider the model selection task in our example \texttt{Workflow}. A user with a penchant for the latest may seek the newest model to showcase to the enterprise, only to be shot down in a presentation by a senior since the latest model makes evaluation against historical benchmarks difficult. In other cases, the risk may be worthwhile if the senior finds something useful for the future from such a model. Another user may want to stay with tried and proven models. For a single agent, serving both such types of user can be a challenge. Two agents with diverse skills can address needs of two different users. These differences among agents can reflect in different qualities as perceived by users and thus different willingness to pay for agent.

\subsubsection{Tenet 6: Low Entry, Exit Barriers for Agents}\hfill

\noindent\underline{Rationale}: Lower barriers to entry foster development of agents (AA6, AA8, PA5). Creating ease of exit for poor performers maintains performance excellence that increases utility for users (UA5, AA6, PA4--5). For entry, planner's gatekeeping role is important, since unlike exit of existing agents, less objective information about agent is expected (PA2, PA5). Moreover, Tenet 6 catalyzes Tenet 5 (AA8, PA5). We refrain from calling out ``free or low'' entry / exit since considerations of sanctity of each enterprise's data / information,  that of switching cost for its users, and proclivity of agents (developers) to overclaim their capability, along with planner's imperfect verifiability of claims, all pose restraints (UA6, PA1, PA6). 

\noindent\underline{Example}: A market mechanism works when an agent with low reward for, say model selection, across a lot of users, falls below a certain threshold and is exited. This also suggests that other qualified agents are ready to take its place, which lower barrier to entry ensures. 

\begin{figure}[ht]
  \centering
  \includegraphics[width=0.8\linewidth]{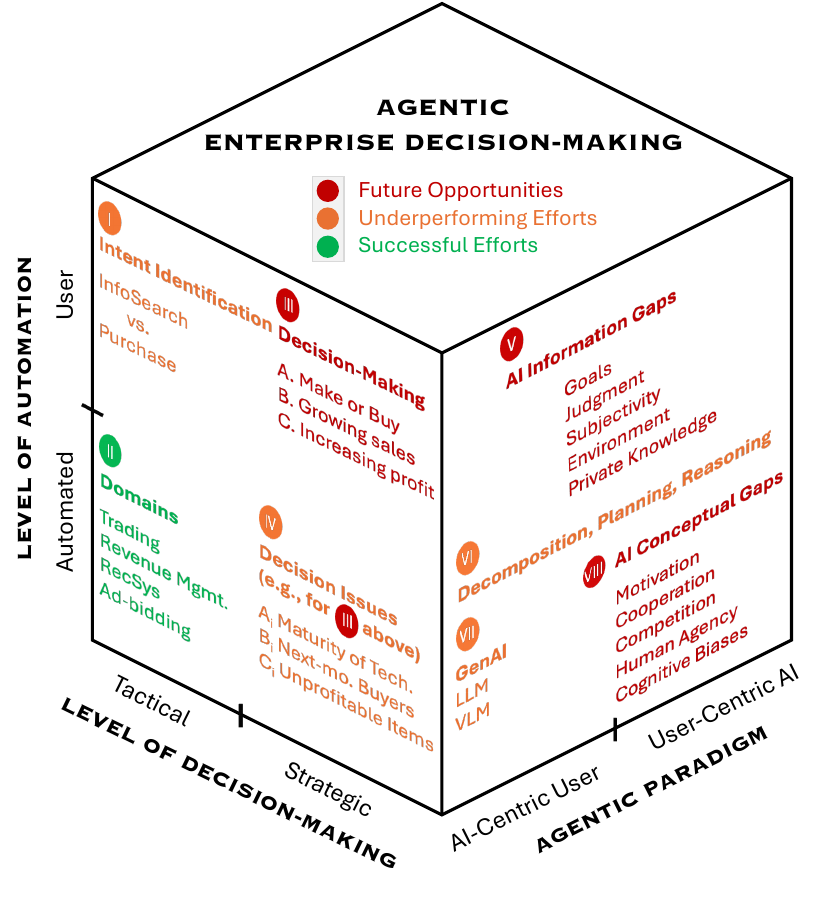}
  \caption{Agentic Enterprise Decision-Making:
  \orangecircleglyph{I}~cautions about intent since it remains difficult to ascertain merely from data without user intervention. \greencircleglyph{II}~highlights the successes of automation for enterprise decision-making in trading in stock market trading; revenue management for dynamic demand systems such as airlines and hotels; recommendation systems for online businesses; and ad-bidding online market mechanism—all with a combination of traditional, ML and AI models. \redcircleglyph{III}~exemplifies gaps in common, big enterprise strategic decisions, none of which is effectively addressable today with any AI. These decisions require enterprise users' breaking them down into atomic decisions issues. \orangecircleglyph{IV}~shows only a single atomic decision issue for each of the three strategic decisions, as broken down by users. \redcircleglyph{V}~shows major information gaps, which are germane for strategic decisions. Next, SOTA models use LLMs in attempts to address atomic issues but fall short. To address atomic issues, \orangecircleglyph{VI}~shows the use of Decomposition, Planning, Reasoning (DPR), de rigueur in strategic decision-making by users. \orangecircleglyph{VII}~shows that LLMs attempt to mimic DPR; but is emblematic of \texttt{AI-Centric~User}~who tries everything to get the LLMs to perform better; yet not rising to a reliable level. Finally, \redcircleglyph{VIII}~highlights the conceptual gaps that are necessary for AI to fill for enterprise decision-making.}
  \Description{An illustration of successful efforts, underperforming efforts, and future opportunities related to Agentic Enterprise Decision-Making across three axes: Level of Automation (Automated versus User), Level of Decision-Making (Tactical versus Strategic), and Agentic Paradigm (AI-Centric User versus User Centric AI).}
  \label{fig:evolution}
\end{figure}

\section{Toward \texttt{User-Centric~AI} for Agentic Enterprise}
Focusing on enterprise decision-making, we propose the type of AI evolution required to pay persistent dividend for companies that use AI. Whether AI is up to the task, or, what form it will take is up for a debate\footnote{https://hbr.org/2024/12/the-irreplaceable-value-of-human-decision-making-in-the-age-of-ai}.  
For the rest of the paper, we refer to Figure~\ref{fig:evolution}. Broadly, decision-making can be classified into Tactical or Strategic; and Automated or Human, where ``automated'' includes human-in-the-loop automated decisions. Our premise is that enterprises' preference is to push more decision-making toward automation (including user-in-the-loop), subject to the decisions outperforming those by users. Favoring their preference is the strong evidence of success in tactical decision-making such as Trading, Revenue Management (e.g., in hotels), Recommendation Systems (RecSYS), Ad-Bidding, etc., where data based automation is de rigueur \greencircleglyph{II}. Tactical decision-making requiring knowledge of user intent \orangecircleglyph{I}, a difficult construct to infer from data alone or due to lack of data~\cite{sheeran2002intention}, still benefits from human involvement. For example, solely based on a sequence of clicks, it is difficult to tell if a new user is only looking for information or wants to make a purchase.

On the strategic decision side, we illustrate with three common examples of enterprise decision problems. For each, the big question is shown in \redcircleglyph{III}, while a corresponding atomic question is shown in \orangecircleglyph{IV}~(only one atomic question among several is shown for each big question). The questions in \orangecircleglyph{IV}~can be answered, in part, as follows by LLMs: 

\begin{enumerate}[nosep, left=0cm]
  \item [\textbf{\textcolor{Maroon}{A.}}] Regarding determining the maturity of a technology, an LLM can respond using its global knowledge gathered from reports and commentaries (about the technology) on the web.
  \item [\textbf{\textcolor{Maroon}{B.}}] Regarding predicting next month's buyers, an LLM's generic global knowledge cannot answer it. In a traditional setting, a business may follow a \texttt{Workflow} (akin to the one in Section~\ref{sec:premise}) to determine the answer. In an LLM setting, a business can provide data about the customer segments it targets and their past purchase behaviors, along with how those segments are defined. The LLM can then suggest a segment definition, though it may not be perfect.
  \item [\textbf{\textcolor{Maroon}{C.}}] Regarding determining unprofitable items, a business may follow a traditional \texttt{Workflow} like \textbf{\textcolor{Maroon}{B}}~'s above. In an LLM setting, the business can feed historical data tables about the different products, unit sales, prices, and costs, for the LLM to provide an answer quickly. However, even through a combination of decomposition, planning, and reasoning operations such as chain of thought and tree of thought~\cite{kumar2023multi,radhakrishnan2023question,nguyen2024interpretable,zhang2024tree,yao2023tree}, LLMs can only \textit{partially} address these questions. 
\end{enumerate}

\noindent These responses obtainable from GenAI are evidently restrictive and consistent with the paradigm of \texttt{AI-Centric~User}; thus, \orangecircleglyph{VII}~shows GenAI as a cautionary tale.

On the other hand, the user dependent big questions are more involved for an LLM to do justice. For instance, in \redcircleglyph{III}~\textbf{\textcolor{Maroon}{A}}, only the business knows its strength and weaknesses in developing the technology and converting it into products. For example, does it have quality human capital to leapfrog the tech beyond what is available or licensed? Does it have the compute and data resources to develop it internally? How long will internal development take? What will it cost to make versus buy, in the short and long term?
This information depends on goals \redcircleglyph{V}~(defensive, to retain current customers, or offensive, to attract new ones); judgment and subjectivity (different executives may have different answers, as these are not objectively measurable); environment (internal technologies and how they fit with the new technology, plus the company's view of its competition); and private knowledge (internal research and confidential insights from C-suite discussions with other companies and tech developers). No LLM has access to this information \redcircleglyph{V}. Moreover, a business may be reluctant to share confidential data with an LLM without an iron-clad leakage prevention or developing its own local GenAI model.

Recent advances in decomposition, planning and reasoning (or, DPR)~\cite{kumar2023multi,radhakrishnan2023question,nguyen2024interpretable,zhang2024tree,yao2023tree} foster hopes of improved automation, with user-in-the-loop. Currently, utility of DPR to enterprises is \emph{limited} \orangecircleglyph{VII}~due to a host of factors including misalignment of goals between AI or Platform and enterprise users, lack of training of finetuning data that capture the complex and involved nature of enterprise decisions, incomplete and imperfect \emph{input} from enterprise users, and dearth of models that incorporate these gaps. The severe gaps \redcircleglyph{V}~with respect to strategic decisions include lack of information about decision makers' goals, lack of data (codified information) about human-judgment, subjectivity (which could also be situational), private knowledge, and  environment (within firm, within industry, etc.). To deal with this ambiguous situation, technology turns to LLMs to fill in the gaps. Users exert effort to extract as much as they can from LLMs, \emph{as if} the yield from LLMs is stymied \textit{only by} users' inability.

Enterprises can benefit from Agents that are endowed with \texttt{User-Centric~AI} \redcircleglyph{VIII}. This kind of AI appreciates human agency, namely, that enterprise users prefer to learn \emph{how to do} their job, instead of being told an answer for a specific question. Users motivation plays a role where motivation is one of two types: prevention or promotion~\cite{higgins1998promotion}, and they are situational and contextual. As illustration, for an analyst with prevention focus, not making an error in presenting results to a large team is more important, while one with a promotion focus may want to surprise the team with a very new analysis and insights. These two analysts differ in their agencies and \textbf{\texttt{User-Centric~AI}} ought to take the difference into account. It is common in enterprises that users cooperate on some work and compete on others~\cite{stigler1974free,hartnell2011organizational}. This has ramifications for Agentic AI, with which all users interact, in that the AI should foster improved cooperation with smoother flow of information among users, but for the latter curtail information flow between users perhaps with sandboxes. Moreover, a firm cooperates with some partner-firms and competes with rival-firms. This understanding, some of which can come from firm-specific information, is crucial for AI to offer decision-making advice and input or make automated decisions on many aspects. Last but not the least, the omnipresence of cognitive biases in managerial judgment and decision-making is well received wisdom~\cite{bazerman2012judgment}. In offering advice to users for decision-making, where users provide input to the AI, detecting such biases in the input affords AI to perform mitigation in advice giving.    

\textit{What kind of Agentic Platform to benefit Enterprises?}  We propose a \textbf{Walled-Garden Platform} with agent autonomy, governed by a market mechanism.  
Enterprises have data and knowledge repositories that need firewall. Without a degree of control exerted over agents, enterprises may find inadequate protection for all their repositories and reduce adoption. At the same time, a centralized planner through which users must access services of agents can be costly in communication (say, planner seeking too many clarifications), poor user experience in not being able to communicate with an agent she specifically needs help from, wrong interpretation of user intent, direction of query to wrong agent, mis-attribution of user's feedback to agents, among others. A walled garden platform with direct communication between users and agents, where users beware of their own actions, planners supervise agents, and agents are rewarded based on--their own claim of performance, planner's observance of their performance and users' feedback--is worthy of strong consideration. Borrowing from the large market mechanism literature in Economics~\cite{myerson2008perspectives,hart1983market}, 
different mechanisms can be designed for different situations. For instance, an ad-bidding platform is a highly successful market mechanism, though it differs significantly from an agentic platform.

\section{Conclusion}
The vision of people and AI working together to perform tasks 
can transition to reality with agents. 
This transition requires taking conceptual notions of how such ecosystems of agents and people will co-exist, and implementing them in specific ways.
In this paper, we propose a framework to reason about the core principles that such ecosystems should exemplify.
We present these as \emph{tenets}, focused on balancing user agency and value as a core principle.
Through these, we posit that future agentic ecosystems can better serve users in enterprises for their decision-making.

Building on this foundation, several key research questions arise: How can we advance locally privacy-preserving agent learning beyond current federated and distributed approaches? Specifically, how can we enable fine-grained user control over which specific information is used for general or targeted learning? Next, what strategies can create a diversity of risk-reward and quality-price trade-offs in both processes and outputs to better meet varied user needs? Lastly, how can we develop simulation environments that test different agent organizations and support diverse user profiles for more effective platform evaluation and refinement?
Addressing these questions will pave the way from \textbf{\texttt{AI-Centric~User}} towards \textbf{\texttt{User-Centric~AI}} in \emph{Agentic Enterprises}.

\bibliographystyle{ACM-Reference-Format}
\bibliography{main}

\end{document}